\documentclass[10pt,twocolumn,letterpaper]{article}

\usepackage[final]{cvpr}      % To produce the REVIEW version
\usepackage{url}            % simple URL typesetting
\usepackage{booktabs}       % professional-quality tables
\usepackage{amsfonts}       % blackboard math symbols
\usepackage{nicefrac}       % compact symbols for 1/2, etc.
\usepackage{microtype}      % microtypography
\usepackage{xcolor}         % colors

\usepackage{adjustbox}
\usepackage{diagbox}
\usepackage{multirow}
\usepackage{makecell}

\usepackage{times}
\usepackage{epsfig}
\usepackage{graphicx}
\usepackage{amsmath}
\usepackage{amssymb}
\usepackage{colortbl}
\usepackage{xspace}

\usepackage{ragged2e}
\usepackage{float}
\usepackage{bbding}
\usepackage{amssymb}  % 需要使用 \checkmark 的话，可能需要导入 amssymb 宏包
\usepackage{bbding}

\definecolor{cvprblue}{rgb}{0.21,0.49,0.74}
\usepackage[pagebackref,breaklinks,colorlinks,citecolor=cvprblue]{hyperref}

\def\paperID{****} % *** Enter the Paper ID here
\def\confName{CVPR}
\def\confYear{2024}

\title{Continual-MAE: Adaptive Distribution Masked Autoencoders for Continual Test-Time Adaptation}

\author{
Jiaming Liu\textsuperscript{\rm 1,2}, 
Ran Xu\textsuperscript{\rm 1}\thanks{Equal contribution, $^{\dagger}$ Project leader, \textsuperscript{\Envelope} Corresponding author.},
Senqiao Yang\textsuperscript{\rm 1,2$^{*}$},
Renrui Zhang \textsuperscript{\rm 3$^{\dagger}$},
Qizhe Zhang\textsuperscript{\rm 1},\\
Zehui Chen \textsuperscript{\rm 4},
Yandong Guo\textsuperscript{\rm 2},
Shanghang Zhang\textsuperscript{\rm 1}~\textsuperscript{\Envelope}
\vspace{0.1cm}\\
\textsuperscript{\rm 1}National Key Laboratory for Multimedia Information Processing, School of Computer Science, \\Peking University
\textsuperscript{\rm 2}AI$^{2}$Robotics   \textsuperscript{\rm 3} MMLab, CUHK  \textsuperscript{\rm 4}University of Science and Technology of China \\
jiamingliu@stu.pku.edu.cn, xu\_ran@bupt.edu.cn, shanghang@pku.edu.cn
}

\begin{document}
\maketitle
\begin{abstract}

Continual Test-Time Adaptation (CTTA) is proposed to migrate a source pre-trained model to continually changing target distributions, addressing real-world dynamism. Existing CTTA methods mainly rely on entropy minimization or teacher-student pseudo-labeling schemes for knowledge extraction in unlabeled target domains. However, dynamic data distributions cause miscalibrated predictions and noisy pseudo-labels in existing self-supervised learning methods, hindering the effective mitigation of error accumulation and catastrophic forgetting problems during the continual adaptation process. To tackle these issues, we propose a continual self-supervised method, Adaptive Distribution Masked Autoencoders (ADMA), which enhances the extraction of target domain knowledge while mitigating the accumulation of distribution shifts. Specifically, we propose a Distribution-aware Masking (DaM) mechanism to adaptively sample masked positions, followed by establishing consistency constraints between the masked target samples and the original target samples. Additionally, for masked tokens, we utilize an efficient decoder to reconstruct a hand-crafted feature descriptor (e.g., Histograms of Oriented Gradients), leveraging its invariant properties to boost task-relevant representations. Through conducting extensive experiments on four widely recognized benchmarks, our proposed method attains state-of-the-art performance in both classification and segmentation CTTA tasks. Our project page: \href{https://sites.google.com/view/continual-mae/home}{https://sites.google.com/view/continual-mae/home}

\end{abstract}    
\section{Introduction}
\label{sec:intro}

\begin{figure}[t]
\includegraphics[width=0.48\textwidth]{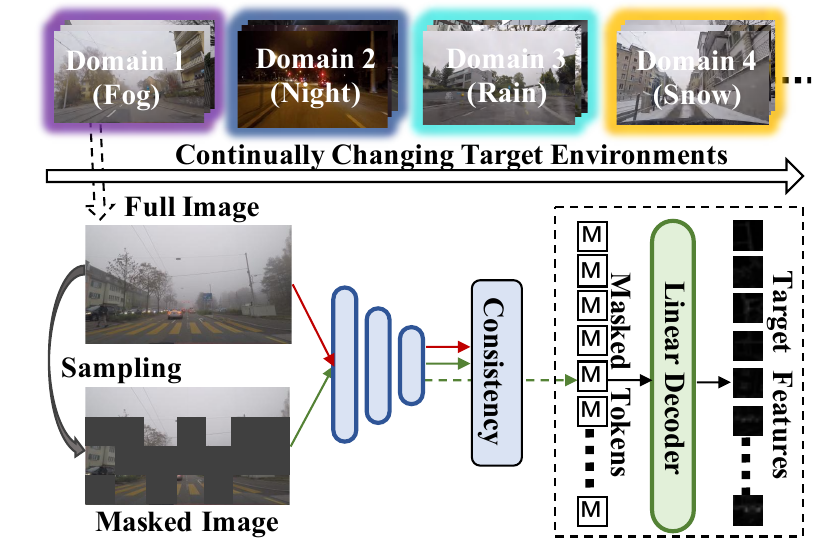}
\vspace{-0.6cm}
\centering
\caption{
In continually changing environments, existing methods \cite{wang2020tent, song2023ecotta} primarily focus on applying entropy minimization to update the normalization layer. However, these approaches are susceptible to miscalibrated predictions, resulting in uncontrollable error accumulation. Alternative mainstream approaches \cite{wang2022continual, gan2022decorate} involve the teacher-student scheme for generating pseudo labels, but noisy pseudo labels limit the model's ability for continuous generalization. In this paper, we propose a novel approach to continual self-supervised learning known as Adaptive Distribution Masked Autoencoders (ADMA). ADMA introduces the mask reconstruction mechanism to enhance the extraction of target domain knowledge while mitigating the domain shift accumulation.
}
\label{fig:intro}
\vspace{-0.4cm}
\end{figure}

Deep Neural Networks (DNNs) have demonstrated impressive performance across various computer vision tasks, including image-level classification \cite{he2016deep,dosovitskiy2020image, liu2021swin}, dense prediction \cite{ren2015faster, zhu2020deformable, xie2021segformer}, and multi-model task \cite{yang2023improved, yang2023lidar, li2023manipllm}, when the test data distribution closely aligns with the training data. 
However, this assumption is frequently challenged in real-world scenarios due to dynamic environments, with deployed models exhibiting insufficient generalization capabilities and performance degradation \cite{sakaridis2021acdc,hendrycks2019benchmarking}. 
Therefore, the problem of continual test-time adaptation (CTTA) has been introduced \cite{wang2022continual}, aiming to adapt a source pre-trained model to continually changing target distributions. Due to privacy and practical considerations, during the continual adaptation process, access to source domain data is not permitted, and each target data can only be accessed once. While the CTTA showcases promising potential applications, it also increases the difficulty of transfer learning, which introduces catastrophic forgetting and error accumulation problems.

Existing methods primarily focus on applying entropy minimization to update batch normalization layer \cite{wang2020tent, gong2022note, niu2022efficient} or a fraction of model parameters \cite{song2023ecotta}, which already leads to a performance improvement in target domains. Nonetheless, due to the continually changing environments, this self-training approach is susceptible to miscalibrated predictions, resulting in uncontrollable error accumulation. 
On the other hand, an alternative mainstream approach involves the teacher-student scheme for generating pseudo-labels in target domains. However, the traditional mean teacher method \cite{tarvainen2017mean} yields noisy pseudo labels in dynamic environments, leading to the accumulation of distribution shifts. While \cite{wang2022continual, liu2023vida, dobler2023robust} utilize the test-time augmentation method to enhance the accuracy of pseudo labels, it limits efficiency during the CTTA process.

In this paper, as shown in Figure \ref{fig:intro}, we introduce a novel approach to continual self-supervised learning called Adaptive Distribution Masked Autoencoders (ADMA). 
Classical masked autoencoders (MAE) \cite{he2022masked} have the potential for various extensions and are becoming dominant in vision representation learning. The selection of reconstruction target and masked positions is particularly crucial during the pretraining process. Nonetheless, reconstructing low-level RGB signals in MAE is considered primitive and redundant, falling short of unlocking the potential of MAE in the context of downstream vision tasks \cite{gao2023mimic, hou2022milan}. To create more effective reconstruction, several methods \cite{hou2022milan, wei2022mvp, wei2022masked, baevski2022data2vec} have explored the utilization of off-the-shelf vision foundation models \cite{oquab2023dinov2, kirillov2023segment} as high-level supervisory signals and designed task-specific mask selection strategies.
Different from previous MAE methods, we make the first attempt to introduce reconstruction techniques to address the continual adaptation problem. This innovation enhances the extraction of target domain knowledge while reducing the accumulation of distribution shifts.

Specifically, we propose a Distribution-aware Masking (DaM) mechanism to distinguish image patches that are target domain-specific from the less significant background patches. The objective is to enhance the quality of the target domain representation, preventing error accumulation and enhancing the efficiency of continuous adaptation. DaM dynamically selects masked positions based on token-wise uncertainty estimation and places learnable masks on token embeddings with substantial domain shifts. Subsequently, it establishes consistency constraints between the network outputs generated from the masked target samples and those from the original target samples. 
Furthermore, for the masked tokens, we employ an efficient decoder to reconstruct hand-crafted feature descriptors, such as Histograms of Oriented Gradients (HOG). In contrast to pixel colors and high-level feature reconstruction, HOG excels at capturing local shapes and appearances, exhibiting partial invariance to geometric and distribution changes \cite{dalal2005histograms, wei2022masked}.
Consequently, we harness its invariant properties to acquire task-relevant representations in target domains, mitigating the impact of domain shifts during continual adaptation and preventing catastrophic forgetting problems.
In summary, our contributions are as follows:

%v1 0506
\begin{itemize}
\item   
We make the first attempt to introduce reconstruction techniques to address the CTTA problem. Our approach, Adaptive Distribution Masked Autoencoders (ADMA), is a novel method for continual self-supervised learning that enhances the extraction of target domain knowledge while mitigating the accumulation of distribution shift.

\item 
In ADMA, we propose a Distribution-aware Masking (DaM) mechanism to adaptively place learnable masks on token embeddings with significant distribution shifts, promoting the quality of target domain representation and improving continual adaptation efficiency.

\item 
For masked tokens, we utilize an efficient decoder to reconstruct histograms of oriented gradients, leveraging its invariant properties to boost task-relevant representations and prevent the catastrophic forgetting problem.

\item 
Our proposed approach surpasses previous state-of-the-art methods, as demonstrated in experiments across four benchmark datasets, covering both classification and segmentation CTTA. Notably, our method attains a promising 87.4\% accuracy in CIFAR10-to-CIFAR10C and 61.8\% mIoU in Cityscapes-to-ACDC scenarios.
\end{itemize}

\begin{figure*}[t]
\centering
% \vspace{-0.1cm}
\includegraphics[width=\linewidth]{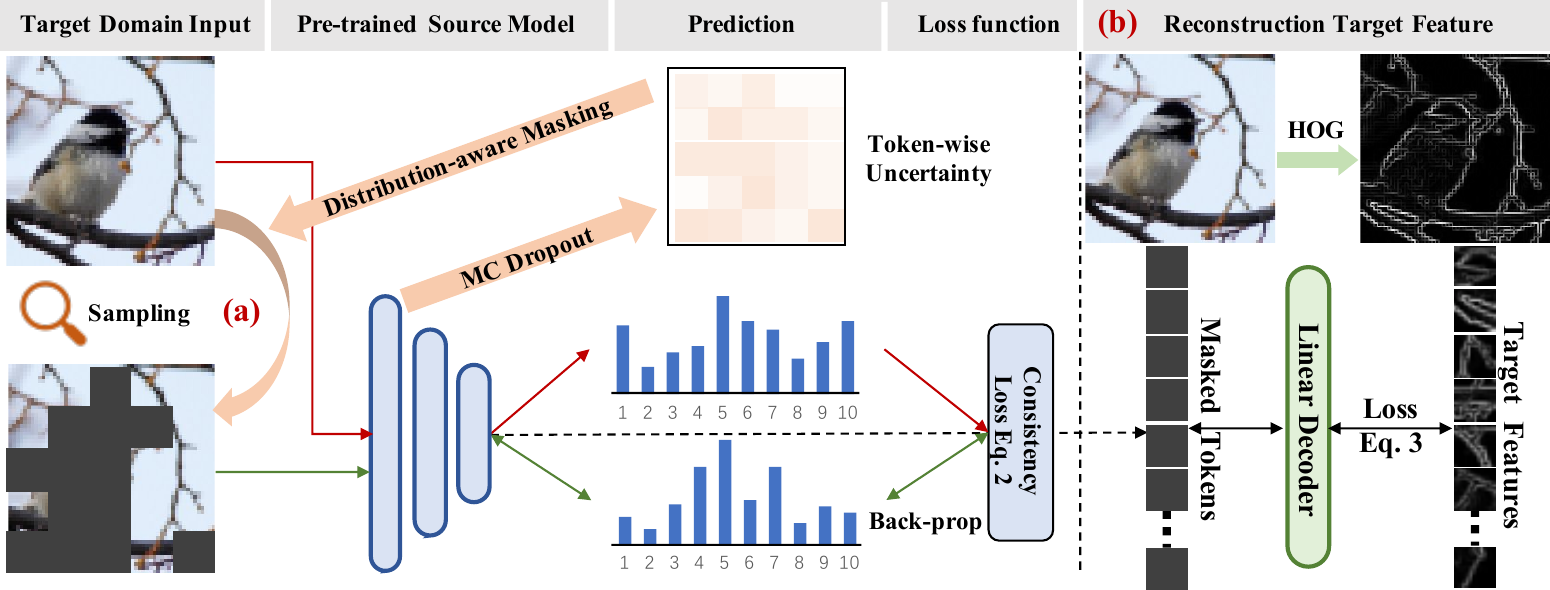}
\vspace{-0.6cm}
\caption{\textbf{The framework of Adaptive Distribution Masked Autoencoders (ADMA).} 
\textcolor{red}{(a)} 
We initiate the process by feeding the original target image into the model to generate features of the complete image. Simultaneously, this step facilitates the estimation of token-wise uncertainty, reflecting the token-wise distribution shift of each target sample, a process detailed in Sec. \ref{sec: DaM}.
Guided by the uncertainty values, we adaptively mask P\% of the image tokens characterized by significant domain shifts, subsequently reintroducing the masked image into the model. In the classification task, the encoder's output embeddings are then fed into the classification heads, constructing a consistency loss (Eq. \ref{eq:con}) between the two predictions. \textcolor{red}{(b)} For the masked tokens, we feed the masked token features into the linear decoder to compute the reconstruction loss (Eq. \ref{eq:rec}). We choose Histograms of Oriented Gradients (HOG) as the reconstruction target due to their invariant properties. Both losses are jointly optimized to address the CTTA problem.
}
\label{fig:framework} 
\vspace{-0.3cm}
\end{figure*}

\section{Related Work}
\subsection{Continual Test-Time Adaptation}

\textbf{Test-time adaptation (TTA)}, also known as source-free domain adaptation~\cite{Boudiafetal2023, Kunduetal2022, ShiqiYangetal2021, zang2023boosting}, is the process of adapting a source model to a target domain distribution without relying on any source domain data. Recent works have delved into techniques such as self-training and entropy regularization to fine-tune the source model~\cite{Liangetal2020, Chenetal2022}. Tent~\cite{DequanWangetal2021} achieves this by updating the training parameters in batch normalization layers through entropy minimization. This approach has served as inspiration for subsequent research efforts in recent works~\cite{niu2023towards, yuan2023robust}, which continue to investigate the robustness of normalization layers.And ~\cite{gandelsman2022test} first attempts to reconstruct RGB images in the TTA task.
\textbf{Continual Test-Time Adaptation (CTTA)} denotes a scenario where the target domain is dynamically changing, introducing additional challenges for conventional TTA methods. The initial approach is presented in~\cite{wang2022continual}, which employs a teacher-student framework to integrate bi-average pseudo labels and stochastic weight reset. Drawing from the insight that mean teacher predictions are often more robust than standard models~\cite{tarvainen2017mean}, a series of mainstream methods~\cite{gan2023decorate, liu2023vida, dobler2023robust} continue this approach to self-supervised learning in CTTA.
Concurrently, existing methods also focus on applying entropy minimization to update normalization layers~\cite{wang2020tent, gong2022note} or a fraction of model parameters~\cite{song2023ecotta, ni2023distribution}. 
However, due to continually changing environments, these self-training approaches are susceptible to miscalibrated predictions and noisy pseudo-labels, resulting in uncontrollable error accumulation.

\subsection{Masked Image Modeling}

Mask-reconstruction-based self-supervised learning has been successful in reducing the reliance on extensive labeled datasets in both Natural Language Processing (NLP) and Computer Vision (CV). The concept was first introduced by BERT \cite{devlin2018bert} in NLP, where a portion of the input word tokens is randomly masked, and the model learns to reconstruct the vocabularies of these masked tokens.
In the field of computer vision (CV), similar techniques have been applied in various works \cite{bao2021beit, he2022masked, xie2022simmim}. These methods involve randomly masking a significant percentage of input image patches. Specifically, BEiT \cite{bao2021beit} was the first to explore Masked Image Modeling (MIM) in vision transformers by reconstructing the vision dictionary derived from DALL-E \cite{ramesh2021zero, ramesh2022hierarchical}. MAE \cite{he2022masked} scaled up MIM to larger models and demonstrated that a simple pixel reconstruction loss can enhance the visual representations of pre-trained models. However, relying solely on low-level RGB signals in MAE is considered rudimentary and limited in unlocking the full potential of MAE in downstream vision tasks.
Subsequently, approaches like MaskFeat \cite{wei2022masked}, data2vec \cite{baevski2022data2vec}, MVP \cite{wei2022mvp}, and MILAN \cite{hou2022milan} have uncovered various high-level signals \cite{zhang2023learning, guo2023joint}, including pre-trained DINO features \cite{oquab2023dinov2}, hand-crafted features \cite{dalal2005histograms}, momentum features \cite{he2020momentum}, and multi-modality features \cite{Radfordetal2021}. Utilizing these high-level signals has been proven to be more effective in extracting contextual information.

\section{Method}
\label{sec:method}

\subsection{Overview}
\textbf{Preliminary.} In Continual Test-Time Adaptation (CTTA), we first pre-train the model $q(y|x)$ using the source domain data $D_{S}={(Y_S, X_S)}$. Subsequently, we adapt $q(y|x)$ to dynamic target domains, denoted as ${D_{T_1}, D_{T_2}, ..., D_{T_n}}$, where $D_{T_i}={{(X_{T_i})}}_{i=1}^{n}$ and $n$ represents the number of continual target datasets. The entire process is restricted from accessing source data and can only utilize each target sample once. Our goal is to continually adapt the pre-trained model to these target domains while mitigating issues such as error accumulation and catastrophic forgetting.

\textbf{Adaptive Distribution Masked Autoencoders} 
Prevalent CTTA approaches primarily focus on applying entropy minimization to update the batch normalization layer \cite{wang2020tent, gong2022note, niu2022efficient} or a fraction of model parameters \cite{song2023ecotta}. However,  due to the dynamic environments, this self-training approach is susceptible to miscalibrated predictions, resulting in uncontrollable error accumulation.
On the other hand, alternative mainstream approaches utilize the teacher-student pseudo-labeling in continual target domains. Nevertheless, the conventional mean teacher method \cite{tarvainen2017mean} increases computational cost and produces noisy pseudo labels in dynamic environments, leading to the accumulation of distribution shifts. While methods such as \cite{wang2022continual, liu2023vida, dobler2023robust} utilize Test-time augmentation methods to enhance the accuracy of pseudo labels, they may limit efficiency during the CTTA process.
Different from prior continual self-supervised approaches, we lead the way in incorporating a masked autoencoder (MAE) to tackle the CTTA problem, abandoning the impact of miscalibrated predictions and the cumbersome teacher-student model.
Our key insight lies in adopting the reconstruction scheme to effectively extract target domain knowledge while mitigating the accumulation of domain shift. 
The overall framework is illustrated in Figure \ref{fig:framework}.
Specifically, We propose a Distribution-aware Masking (DaM) mechanism, detailed in Sec. \ref{sec: DaM}, which adaptively places learnable masks on token embeddings with substantial distribution shifts. 
By establishing consistency constraints between the masked input and the original input, DaM significantly enhances the understanding of target domain knowledge and mitigates the challenge of error accumulation. 
For masked tokens, we adopt a linear decoder to reconstruct Histograms of Oriented Gradients, leveraging its invariant properties to acquire task-relevant representations while averting the introduction of target domain shifts, as elaborated in Sec. \ref{sec: HOG}. This reconstruction method serves as a preventive measure, avoiding catastrophic forgetting in continual adaptation. We provide the intuitive explanation and justification in Sec. \ref{sec: justify}.

\subsection{Distribution-aware Masking} 
\label{sec: DaM}

To establish masked image reconstruction as a meaningful pretext task, previous studies have commonly applied an aggressive masking approach by randomly masking a substantial portion of input image patches \cite{he2022masked, wei2022masked}. This strategy, however, introduces a potential drawback: the remaining visible patches may predominantly comprise background information, potentially lacking the crucial cues essential for reconstructing foreground details \cite{hou2022milan}.
In our framework, the precise selection of masked positions is crucial due to significant distribution shifts in each target sample, impacting the representational capacity of visible patches. Furthermore, in the CTTA task, unlike traditional MAE pre-training, each sample is encountered only once, demanding high efficiency in the reconstruction process.

To this end, as shown in Figure \ref{fig:framework} (a), we introduce a Distribution-aware Masking (DaM) strategy, enabling a thoughtful selection of tokens to mask in dynamic environments. The key concept involves choosing tokens with substantial domain shifts for masking, ensuring that the preserved visible tokens exhibit relatively fewer domain shifts while providing reliable semantic knowledge through the model encoder.
To quantify token-wise distribution shifts, we draw inspiration from \cite{zhang2023efficient, zhang2023unimodal, ovadia2019can, roy2022uncertainty} and introduce a method for token-wise uncertainty estimation. Specifically, we employ the MC Dropout \cite{gal2016dropout}, enabling multiple forward propagations to obtain $m$ (e.g., $m = 10$) sets of features for each token. Subsequently, we calculate the uncertainty value $\mathcal{U}(x)$ for a given token $x_j$, as formulated below:
\begin{equation}
% \vspace{-0.2cm}
\mathcal{U} (x_j) =  \left( \frac{1}{m} \sum_{i=1}^m \|f_i(x_j) - \mu \|^2 \right) ^{\frac{1}{2}}
\label{eq:mc}
% \vspace{-0.1cm}
\end{equation}
Where $f_i(x_j)$ is the feature value of the token $x_j$ in the $i^{th}$ forward propagation, and $\mu$ is the average value of the token feature over $m$ forward propagations. To calculate the feature value within a token, we utilize average pooling, reducing the token's dimension from $1 \times 768$ to $1 \times 1$.  
Note that we only apply MC Dropout to the linear layer within the FFN layer in the first Transformer block. We conduct m forward propagations in the local FFN layer, calculating the uncertainty value does not significantly increase computational cost.
In this manner, we obtain the uncertainty value for each token. After sorting, we select the top P\% (e.g., 50\%) of tokens with the highest uncertainty for masking.

We input the masked image into the model, leveraging the remaining contextual clues to reconstruct the class label. Subsequently, we establish consistency constraints between the network outputs ($\hat{y}(c), y(c)$) generated from the masked target samples and those from the original target samples. The formulation of the consistency loss is as follows:
\begin{equation}
 \mathcal{L}_{con}(x) = - 
 \frac{1}{C}
 \sum_c^C y(c) \log \hat{y}(c)
\label{eq:con}
\end{equation}
Where $\hat{y}$ is the output from the masked image, $C$ means the number of categories. 
Through DaM and consistency constraints, we can mask a substantial amount of distribution shift, learning the target domain contextual knowledge while avoiding domain shift accumulation.

\subsection{Reconstruction Target Feature} 
\label{sec: HOG}
The reconstruction target plays an important role in masked image modeling, exerting a direct influence on the learning of feature representations.
Previous MAE methods typically opt for either low-level RGB information \cite{he2022masked} or high-level semantic information \cite{hou2022milan, gao2023mimic} as the reconstruction target. However, in the CTTA tasks, reconstructing the RGB signal of the target domain introduces inherent domain shifts in the reconstruction process. Similarly, when reconstructing the semantic features of the target domain, such as features from CLIP \cite{Radfordetal2021}, the absence of any operation to domain transfer during the feature extraction process means that this approach still fails to alleviate the domain shift. These statements are demonstrated in Sec. \ref{sec: abla}. Therefore, in the face of continual distribution shifts, the choice of the reconstruction target gains greater importance. 

\begin{figure}[t]
\includegraphics[width=0.48\textwidth]{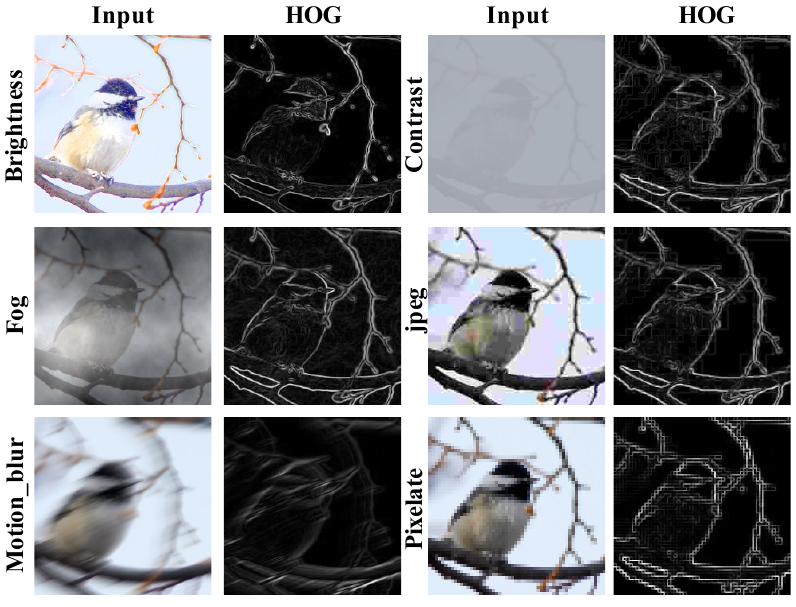}
\vspace{-0.6cm}
\centering
\caption{
The visualization of HOG features in various target domain distributions (ImageNet-C \cite{hendrycks2019benchmarking}).
}
\label{fig:hog}
\vspace{-0.6cm}
\end{figure}

Inspired by \cite{wei2022masked} in model pretraining, we introduce Histograms of Oriented Gradients (HOG) reconstruction in CTTA tasks, showcasing notable benefits in the continual adaptation process.  HOG is a feature descriptor that delineates the distribution of gradient orientations or edge directions within a localized subregion \cite{dalal2005histograms}. Using HOG as the reconstruction target in CTTA offers two advantages: 1) its inherent ability to capture local shapes and appearances ensures invariance to geometric changes, and 2) the absorption of brightness through image gradients and local contrast normalization provides invariance to varying environments and weather conditions. 
As illustrated in Figure \ref{fig:hog}, the visualization of HOG under various target domain distributions presents similar feature representations, clearly indicating their invariant properties.

To obtain HOG features, we employ a two-channel convolution that generates gradients along the x and y axes, followed by histogramming and normalization. Following \cite{wei2022masked}, we configure the orientation bins to be 9, spatial cell size to be $8 \times 8$, and channels to be 3. Consequently, the HOG feature $F_{HOG} \in \mathbb{R}^{3 \times 9 \times H/8 \times W/8}$, where H and W represent the height and width of the input image. 
After obtaining the HOG features, we employ a linear layer to project learnable masked tokens to the same dimension as $F_{HOG}$, minimizing the L2 distance between the HOG prediction $P_{HOG}$ and HOG label $F_{HOG}$ of the masked token positions. The reconstruction loss is formulated as: 

\begin{equation} \label{eq:rec}
    % \mathcal{L}_{rec} = \frac{1}{N}\sum_{n}^N \left\| P_{HOG}  - G_{HOG}  \right\|_{2}^{2} ,
    \mathcal{L}_{rec} = \left\| P_{HOG}  - F_{HOG}  \right\|_{2}^{2}
\end{equation}
Through HOG reconstruction, we harness its invariant properties to extract task-relevant knowledge in the CTTA problem. This enables the model to concentrate more on the task at hand and mitigates the impact of domain shift, consequently reducing the risk of catastrophic forgetting.

\subsection{Optimization Objective}

In the ongoing adaptation process, we update the model by incorporating the total loss formulated as Eq. \ref{eq:loss}, where $\lambda = 0.5$ serves as a balancing factor for the loss values.
\begin{equation} \label{eq:loss}
    \mathcal{L}_{all} = \mathcal{L}_{con} + \lambda \times \mathcal{L}_{rec}
\end{equation}
\section{Experiments}
\label{sec:Experiments}

\begin{table*}[t]
% \vspace{-0.1cm}
\centering
\small
\setlength\tabcolsep{2pt}%调列距
\begin{adjustbox}{width=1\linewidth,center=\linewidth}
\begin{tabular}{c|c|ccccccccccccccc|cc}
\hline
% Time & \multicolumn{15}{l|}{$t\xrightarrow{\hspace*{13.5cm}}$}& \\ \hline
 Method & REF &
 \rotatebox[origin=c]{50}{Gaussian} & \rotatebox[origin=c]{50}{shot} & \rotatebox[origin=c]{50}{impulse} & \rotatebox[origin=c]{50}{defocus} & \rotatebox[origin=c]{50}{glass} & \rotatebox[origin=c]{50}{motion} & \rotatebox[origin=c]{50}{zoom} & \rotatebox[origin=c]{50}{snow} & \rotatebox[origin=c]{50}{frost} & \rotatebox[origin=c]{50}{fog}  & \rotatebox[origin=c]{50}{brightness} & \rotatebox[origin=c]{50}{contrast} & \rotatebox[origin=c]{50}{elastic\_trans} & \rotatebox[origin=c]{50}{pixelate} & \rotatebox[origin=c]{50}{jpeg}
& Mean$\downarrow$ & Gain\\\hline
Source \cite{dosovitskiy2020image}& ICLR2021&60.1&53.2&38.3&19.9&35.5&22.6&18.6&12.1&12.7&22.8&5.3&49.7&23.6&24.7&23.1&28.2&0.0\\
Pseudo-label \cite{Leeetal2013}& ICML2013&59.8&52.5&37.2&19.8&35.2&21.8&17.6&11.6&12.3&20.7&5.0&41.7&21.5&25.2&22.1&26.9&+1.3\\
TENT-continual \cite{DequanWangetal2021}& ICLR2021&57.7&56.3&29.4&16.2&35.3&16.2&12.4&11.0&11.6&14.9&4.7&22.5&15.9&29.1&19.5&23.5&+4.7\\
CoTTA \cite{wang2022continual}& CVPR2022&58.7&51.3&33.0&20.1&34.8&20&15.2&11.1&11.3&18.5&4.0&34.7&18.8&19.0&17.9&24.6&+3.6\\
VDP \cite{gan2023decorate} & AAAI2023&57.5&49.5&31.7&21.3&35.1&19.6&15.1&10.8&10.3&18.1&4.0&27.5&18.4&22.5&19.9&24.1&+4.1\\
ViDA \cite{liu2023vida}& ICLR2024 &52.9&47.9&19.4&11.4&31.3&\textbf{13.3}&\textbf{7.6}&7.6&9.9&12.5&\textbf{3.8}&26.3&14.4&33.9&18.2&20.7&+7.5 \\
\textbf{Ours} & \textbf{Proposed} & \textbf{30.6} & \textbf{18.9} & \textbf{11.5} & \textbf{10.4} & \textbf{22.5} & 13.9 & 9.8 & \textbf{6.6} & \textbf{6.5} & \textbf{8.8} & 4.0 & \textbf{8.5} & \textbf{12.7} & \textbf{9.2} & \textbf{14.4}
&\textbf{12.6}& +\textbf{15.6}\\
\hline
\end{tabular}
\end{adjustbox}
\vspace{-0.3cm}
\caption{\label{tab: cifar10}Classification error rate(\%) for CIFAR10-to-CIAFAR10C online CTTA task. Mean(\%) denotes the average error rate across 15 target domains. Gain(\%) represents the percentage of improvement in model accuracy compared with the source method.}
\vspace{-0.2cm}
\end{table*}

\begin{table*}[t]
% \vspace{-0.1cm}
\small
\centering
\setlength\tabcolsep{2pt}%调列距
\begin{adjustbox}{width=1\linewidth,center=\linewidth}
\begin{tabular}{c|c|ccccccccccccccc|cc}
\hline
% Time & \multicolumn{15}{l|}{$t\xrightarrow{\hspace*{13.5cm}}$}& \\ \hline
  Method & REF &
 \rotatebox[origin=c]{50}{Gaussian} & \rotatebox[origin=c]{50}{shot} & \rotatebox[origin=c]{50}{impulse} & \rotatebox[origin=c]{50}{defocus} & \rotatebox[origin=c]{50}{glass} & \rotatebox[origin=c]{50}{motion} & \rotatebox[origin=c]{50}{zoom} & \rotatebox[origin=c]{50}{snow} & \rotatebox[origin=c]{50}{frost} & \rotatebox[origin=c]{50}{fog}  & \rotatebox[origin=c]{50}{brightness} & \rotatebox[origin=c]{50}{contrast} & \rotatebox[origin=c]{50}{elastic\_trans} & \rotatebox[origin=c]{50}{pixelate} & \rotatebox[origin=c]{50}{jpeg}
& Mean$\downarrow$ & Gain\\\hline
Source \cite{dosovitskiy2020image}&ICLR2021&55.0&51.5&26.9&24.0&60.5&29.0&21.4&21.1&25.0&35.2&11.8&34.8&43.2&56.0&35.9&35.4&0.0\\
Pseudo-label \cite{Leeetal2013}& ICML2013&53.8&48.9&25.4&23.0&58.7&27.3&19.6&20.6&23.4&31.3&11.8&28.4&39.6&52.3&33.9&33.2&+2.2\\
TENT-continual  \cite{DequanWangetal2021} &ICLR2021&53.0&47.0&24.6&22.3&58.5&26.5&19.0&21.0&23.0&30.1&11.8&25.2&39.0&47.1&33.3&32.1&+3.3\\
CoTTA \cite{wang2022continual}& CVPR2022&55.0&51.3&25.8&24.1&59.2&28.9&21.4&21.0&24.7&34.9&11.7&31.7&40.4&55.7&35.6&34.8&+0.6\\
VDP \cite{gan2023decorate} & AAAI2023&54.8&51.2&25.6&24.2&59.1&28.8&21.2&20.5&23.3&33.8&\textbf{7.5}&\textbf{11.7}&32.0&51.7&35.2&32.0&+3.4\\
ViDA  \cite{liu2023vida} & ICLR2024 &50.1 & 40.7 & 22.0 & \textbf{21.2} & 45.2 & \textbf{21.6} & \textbf{16.5} & \textbf{17.9} & \textbf{16.6} & 25.6 & 11.5 & 29.0 & \textbf{29.6} & \textbf{34.7} & \textbf{27.1} & 27.3 & +8.1\\
\textbf{Ours} & \textbf{Proposed} & \textbf{48.6} & \textbf{30.7} & \textbf{18.5} & 21.3 & \textbf{38.4} & 22.2 & 17.5 & 19.3 & 18.0 & \textbf{24.8} & 13.1 & 27.8 & 31.4 & 35.5 & 29.5& \textbf{26.4} &\textbf{+9.0} \\
\hline
\end{tabular}
\end{adjustbox}
\vspace{-0.3cm}
\caption{\label{tab: cifar100}Classification error rate(\%) for CIFAR100-to-CIAFAR100C online CTTA task.}
\vspace{-0.3cm}
\end{table*}

In Sec. \ref{sec: 4.1}, we present experiments and implementation details. Sec. \ref{sec: 4.2} and Sec. \ref{sec: 4.3} provide a comparative analysis of our approach against previous methods in classification and semantic segmentation CTTA tasks. Furthermore, we conduct a comprehensive ablation study in Sec. \ref{sec: abla}. Due to page constraints, additional quantitative and qualitative analyses are available in Appendices A and B, respectively.

\subsection{Experiments Details}
\label{sec: 4.1}

\textbf{Datasets.} Our method undergoes evaluation on three classification CTTA benchmarks, which encompass CIFAR10-to-CIFAR10C, CIFAR100-to-CIFAR100C \cite{krizhevsky2009learning}, and ImageNet-to-ImageNet-C \cite{hendrycks2019benchmarking}. In the segmentation CTTA, following the definition by \cite{wang2022continual, yang2023exploring}, we conduct assessments on the Cityscapes-to-ACDC, using the Cityscapes \cite{yang2023exploring} as the source domain and the ACDC \cite{sakaridis2021acdc} as the target domain.

\textbf{Task setting.} Following the task setting outlined in \cite{wang2022continual, liu2023vida}, in the classification CTTA tasks, we utilize a sequential adaptation process. The pre-trained source model adapts to each of the fifteen target domains characterized by the largest corruption severity (level 5). Online prediction results are immediately assessed after processing the input.
For the segmentation CTTA task, we use the ACDC \cite{sakaridis2021acdc} dataset to represent the target domain, which includes images captured under four distinct unobserved visual conditions: Fog, Night, Rain, and Snow. To simulate continuous environmental changes resembling real-world scenarios, we cyclically iterate through the same sequence of target domains (Fog → Night → Rain → Snow) multiple times.

\textbf{Implementation Details.} In our CTTA experiments, to ensure consistency and fair comparisons, we follow the implementation details as proposed in the prior CTTA works \cite{wang2022continual,liu2023vida}. For the classification CTTA tasks, we employ the ViT-base  \cite{dosovitskiy2020image}  as our backbone model. We resize the input images to 384$\times$384 for CIFAR10C, CIFAR100C benchmark, and 224$\times$224 resolution for ImageNet-C benchmark. In the case of segmentation CTTA task, we employ the Segformer-B5 \cite{xie2021segformer} pre-trained on the Cityscapes dataset as our source model. We down-sample the original input images resolution from 1920$\times$1080 to 960$\times$540. We use Adam \cite{kingma2014adam} with $(\beta_{1}, \beta_{2}) = (0.9, 0.999)$ as the optimizer. Different learning rates are assigned for different CTTA tasks and backbone models, such as we use 1e-5 for ViT on CIFAR10C and CIFAR100C, 1e-3 for ViT on ImageNetC, and 3e-4 for Segformer on ACDC.
In the masking process, following the approach in MIC \cite{hoyer2022mic}, we use [MASK] tokens initialized with all zeros to replace 50\% of tokens that are embedded as patches. These [MASK] tokens are shared learnable embeddings that indicate masked patches. The reconstruction decoder is a randomly initialized linear layer used to project the output tokens associated with masked patches to HOG features. The parameters of mask tokens and projection layer are optimized in parallel with other parameters. To achieve better results, we inject ViDA, as proposed by previous SOTA work \cite{liu2023vida}, into the model and utilize our proposed self-supervised method to update.

\subsection{Classification CTTA Tasks}
\label{sec: 4.2}

\textbf{CIFAR10-to-CIFAR10C \& CIFAR100-to-CIFAR100C.} The source model is trained through supervised learning on the CIFAR10 or CIFAR100 dataset. During testing, we apply CTTA to the CIFAR10C or CIFAR100C dataset, which contains fifteen corruption types continuously fed into the model in a specific order.
In the CIFAR10-to-CIFAR10C scenario, as shown in Table \ref{tab: cifar10}, the average classification error of the source model reaches 28.2\% when directly testing on CIFAR10C. However, our method significantly reduces the error to 12.6\%. Compared to other continual self-supervised methods, our approach outperforms them by 10.9\% and 12.0\% compared to the previous entropy minimization method (TENT \cite{DequanWangetal2021}) and the teacher-student method (CoTTA \cite{wang2022continual}), demonstrating significant potential in addressing the CTTA problem. To be mentioned, our Adaptive Distribution Masked Autoencoders (ADMA) demonstrate outstanding performance across 12 out of the 15 corruption types, validating the robustness of our method in the continual adaptation process. Note that, for TENT, we implement entropy minimization to update the Layer Normalization layers in the transformer instead of BN.

We expand our evaluation to the CIFAR100-to-CIFAR100C scenario, as depicted in Table \ref{tab: cifar100}, encompassing a more extensive range of categories within each domain. Our approach outperforms the previous entropy minimization and teacher-student pseudo-labeling methods by 5.7\% and 8.4\%, respectively, exhibiting a consistent trend with the aforementioned CTTA experiments. Therefore, the results validate the universality of our method, unaffected by the number of categories, and it can efficiently mitigate error accumulation and catastrophic forgetting problems.

\begin{table}[t]
\begin{center}
% \vspace{-0.2cm}
\centering
    \setlength\tabcolsep{0.12cm}%调列距
    \small
    % \setlength\tabcolsep{1pt}
    % \begin{adjustbox}{width=1\linewidth,center=\linewidth}
    \begin{tabular}{c|c|ccccc}
    \hline
    Target&Method & Source& Tent& CoTTA & VDP & \textbf{Ours}\\\hline
    \multirow{2}{*}{ImageNet-C}&Mean$\downarrow$& 55.8&51.0&54.8& 50.0&\textbf{42.5}\\
    &Gain&0.0 &+4.8&+1.0&+5.8 &\textbf{+13.3}\\\hline
    \end{tabular}
    % \end{adjustbox}
    % \vspace{-0.2cm}
% \hspace{0.02\textwidth}
\end{center}
\vspace{-0.5cm}
\caption{\label{tab: imagenet} 
    Average error rate (\%) for the ImageNet-to-ImageNet-C CTTA. The fine-grained performances are shown in Appendix A.}
\vspace{-0.3cm}
\end{table}

\begin{table*}[t]
% \vspace{-0.1cm}
% \vspace{-2cm}
\centering
% \vspace{-0.3cm}
\setlength\tabcolsep{2pt}%调列距
\begin{adjustbox}{width=1\linewidth,center=\linewidth}
\begin{tabular}{c|c|ccccc|ccccc|ccccc|c|c }
\hline

\multicolumn{2}{c|}{Time}     & \multicolumn{15}{c}{$t$ \makebox[10cm]{\rightarrowfill} }                                                                              \\ \hline
\multicolumn{2}{c|}{Round}          & \multicolumn{5}{c|}{1}    & \multicolumn{5}{c|}{2}     & \multicolumn{5}{c|}{3}  & \multirow{2}{*}{Mean$\uparrow$}   & \multirow{2}{*}{Gain}  \\ \cline{1-17}
Method & REF & Fog & Night & Rain & Snow & Mean$\uparrow$ & Fog & Night & Rain & Snow  & Mean$\uparrow$ & Fog & Night & Rain & Snow & Mean$\uparrow$ & \\ \hline
Source \cite{xie2021segformer}& ICLR2021&69.1&40.3&59.7&57.8&56.7&69.1&40.3&59.7& 	57.8&56.7&69.1&40.3&59.7& 57.8&56.7&56.7&/\\ 
TENT \cite{wang2020tent}&ICLR2021 &69.0&40.2&60.1&57.3&56.7&68.3&39.0&60.1& 	56.3&55.9&67.5&37.8&59.6&55.0&55.0&55.7&-1.0\\ 
CoTTA \cite{wang2022continual}& CVPR2022 &70.9&41.2&62.4&59.7&58.6&70.9&41.1&62.6& 	59.7&58.6&70.9&41.0&62.7&59.7&58.6&58.6&+1.9\\
SVDP  \cite{yang2023exploring}&AAAI2024 &\textbf{72.1}&44.0&65.2&63.0&61.1&\textbf{72.2}&44.5&65.9&\textbf{63.5}&61.5&72.1&44.2&65.6&\textbf{63.6}&61.4&61.3&+4.6\\ 
\textbf{Ours} & \textbf{Proposed} & 71.9& \textbf{44.6}& \textbf{67.4}& \textbf{63.2}& \textbf{61.8}& 71.7& \textbf{44.9}& \textbf{66.5}& 63.1& \textbf{61.6} 
 & \textbf{72.3}& \textbf{45.4}& \textbf{67.1}& 63.1& \textbf{62.0} & \textbf{61.8} 
 & +\textbf{5.1}\\\hline
 \hline
\end{tabular}
\end{adjustbox}
\vspace{-0.3cm}
\caption{\label{tab: ACDC} \textbf{Performance comparison for Cityscape-to-ACDC CTTA.} We sequentially repeat the same sequence of target domains three times. Mean(\%) is the average score of mIoU. Gain(\%) represents the improvement of mIoU compared with the source method.}
\vspace{-0.3cm}
\end{table*}

\textbf{ImageNet-to-ImageNet-C.} In order to comprehensively evaluate the effectiveness of our method, we conduct experiments on the ImageNet-to-ImageNet-C scenario. The source model is pre-trained on the ImageNet. 
As indicated in Table \ref{tab: imagenet}, due to the large number of categories in the ImageNet-C, previous entropy minimization (TENT \cite{DequanWangetal2021}) and the teacher-student approach (CoTTA \cite{wang2022continual}) only achieve error rates of 51.0\% and 54.8\%, respectively.
In contrast, our proposed method achieves the best performance, showcasing a significant reduction in classification error rates to 42.5\%. This outcome further demonstrates the effectiveness of our method, enhancing the feature representation of the target domain without succumbing to domain shift accumulation. 

\subsection{Semantic Segmentation CTTA Task}
\label{sec: 4.3}

\textbf{Cityscapes-to-ACDC.} We validate the effectiveness of our approach in the more challenging segmentation CTTA task by adapting the pre-trained Segformer model from the Cityscapes dataset to the ACDC \cite{sakaridis2021acdc} dataset, as depicted in Table \ref{tab: ACDC}. 
To ensure the reliability of the model's pixel-level outputs, we adopt the update strategy from previous works \cite{wang2022continual,liu2023vida}, such as utilizing multi-scale augmentation.
Our proposed method exhibits a notable improvement, achieving a 5.1\% increase in mIoU over the source model, thereby confirming its efficacy in dense prediction CTTA tasks. Moreover, our method outperforms the previous entropy minimization method (TENT \cite{DequanWangetal2021}) and the teacher-student method (CoTTA \cite{wang2022continual}) by 6.1\% and 3.2\%, respectively.
It is worth highlighting the stability of our method in comparison to others. While TENT \cite{DequanWangetal2021} reduces 1.7\% performance from the first to the third round of experiments, CoTTA \cite{wang2022continual} maintains consistent results between these rounds. In contrast, our method demonstrates a 0.2\% increase in mIoU during the third round. This observation underscores the effectiveness of our approach in extracting target domain knowledge through efficient mask modeling. Additionally, our reconstruction scheme ensures task-relevant feature representation, mitigating catastrophic forgetting.

\subsection{Ablation Study}
\label{sec: abla}
%components effectiveness

\begin{table}[t]
\centering
\setlength\tabcolsep{8pt}%调列距
% \begin{adjustbox}{width=1\linewidth,center=\linewidth}
\small
\begin{tabular}{c|ccc|c|c}
\hline
% Time & \multicolumn{15}{l|}{$t\xrightarrow{\hspace*{13.5cm}}$}& \\ \hline
 &Random& DaM &HOG
& Mean$\downarrow$ &Gain \\\hline
Ex0& - & - & - &28.2&/\\
Ex1& \checkmark & -&- &17.1& +11.1\\
Ex2& - &\checkmark &- &14.4& +13.8\\
Ex3& \checkmark & -&\checkmark &15.8& +12.4\\
Ex4&-  & \checkmark&\checkmark &\textbf{12.6} & +\textbf{15.6} \\
\hline
\end{tabular}
% \end{adjustbox}
\vspace{-0.3cm}
\caption{Average error rate(\%) for CIFAR10-to-CIFAR10C online CTTA task. Random, DaM, and HOG represent the random masking strategy, our proposed Distribution-aware Masking mechanism, and our introduced HOG reconstruction, respectively.}
% \vspace{-0.1cm}
\label{tab: abla}
\end{table}

\begin{table}[t]
\begin{center}
% \vspace{-0.2cm}
\centering
    \setlength\tabcolsep{0.3cm}%调列距
    \small
    % \setlength\tabcolsep{1pt}
    % \begin{adjustbox}{width=1\linewidth,center=\linewidth}
    \begin{tabular}{c|cccc}
    \hline
    Target& RGB & CLIP & Mean-Teacher & HOG\\\hline
     Error rate &49.2&48.5&48.1&\textbf{43.6}\\
    \hline
    Target&SIFT  &Sobel& Laplacian& ORB\\\hline
    Error rate &50.4&48.2&48.7&50.0\\
    \hline
    \end{tabular}
    % \end{adjustbox}
    % \vspace{-0.2cm}
% \hspace{0.02\textwidth}
\end{center}
\vspace{-0.6cm}
\caption{\label{tab: recon} 
    Average error rate(\%) for ImageNet-to-ImageNet-C. RGB, CLIP, Mean-Teacher, SIFT \cite{lowe1999object}, ORB, edge detectors(Sobel and Laplacian), and HOG are different reconstruction target.}
\vspace{-0.3cm}
\end{table}

\textbf{Effectiveness of each component.} 
We initially conduct a series of ablation experiments on CIFAR10-to-CIFAR10C to assess the contributions of different components in our approach. As shown in Table \ref{tab: abla}, the first set of experiments (Ex1) involve randomly masking a portion of patches from the input image and establishing consistency constraints between the model outputs generated from the masked target samples and those from the original target samples. This directly led to an 11.1\% reduction in the error rate compared to the source method (Ex0), indicating that the masking strategy enhances target domain knowledge extraction in CTTA.
In the second set of experiments (Ex2), we replace the random masking strategy with our proposed Distribution-aware Masking (DaM) mechanism. Ex2 further reduced the error rate by 2.7\%, validating that DaM helps the model more efficiently understand the target domain distribution. In the subsequent experiments (Ex3 and Ex4), we introduced the Histogram of Oriented Gradients (HOG) reconstruction scheme. For Ex4, HOG reconstruction contributed to a 1.8\% accuracy improvement than Ex2. 
When integrating HOG reconstruction into our proposed DaM, a final classification error rate of 12.6\% is achieved, leading to an overall performance improvement of 15.6\%. These results confirm that leveraging HOG reconstruction during the continual adaptation process assists the model in learning task-relevant knowledge under the presence of domain shifts.

\textbf{Reconstruction target selection.} Another set of ablation experiments aims to assess the impact of the reconstruction target. This includes the original RGB pixel, high-level CLIP features, Mean-Teacher features, as well as SIFT, ORB, Sobel, Laplacian, and HOG features. Since the CIFAR datasets have an input size of $32\times32$ with limited RGB pixel information, these ablation experiments are conducted in the context of the ImageNet-to-ImageNet-C. As shown in Table \ref{tab: recon}, our introduced HOG reconstruction outperforms other reconstruction targets by a significant margin. Specifically, compared with classical image pixel reconstruction, our method achieves a 5.6\% improvement, demonstrating that reconstructing the RGB signal of the target domain introduces inherent domain shifts. In contrast to reconstructing the CLIP feature, our method outperforms it by 4.9\%, while also significantly reducing computational costs as it does not require an additional model. The results validate that features directly extracted by the large-scale model still introduce domain shifts. Lastly, while reconstructing features of the Mean-Teacher model is slightly better (by 0.4\%) than reconstructing CLIP features, it is still 4.5\% lower than our method. In conclusion, reconstructing HOG and leveraging its invariant properties can boost task-relevant representations and avoid domain shift accumulation in CTTA.

\section{Discussion and Justification}
\label{sec: justify}
%CAM vis

\begin{figure}[t]
\includegraphics[width=0.47\textwidth]{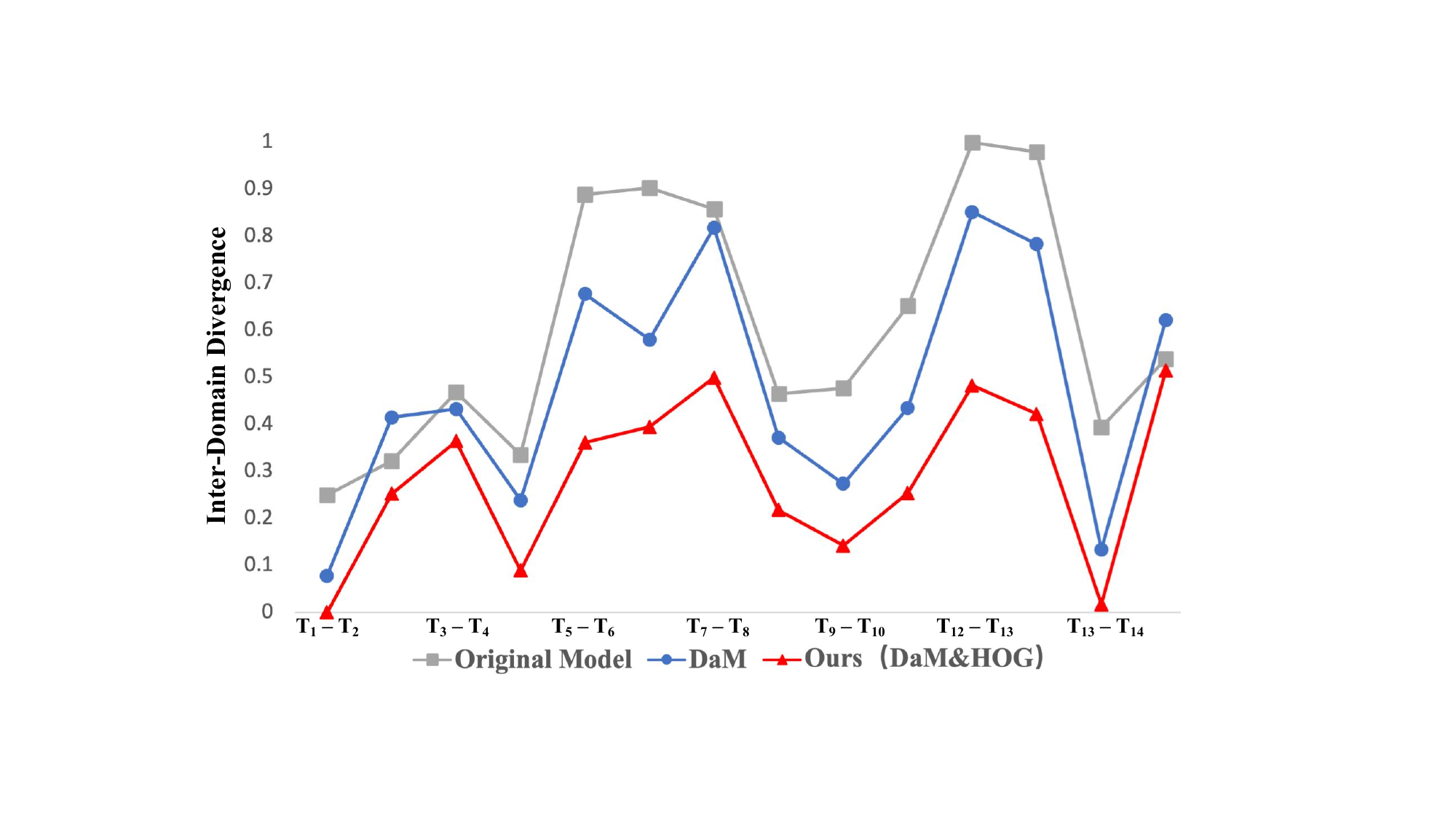}
\vspace{-0.1cm}
\centering
\caption{
The inter-domain divergency. $T_{1}$ to $T_{15}$ represent the 15 target domains in CIFAR-10C, listed in sequential order.
}
\label{fig: div}
\vspace{-0.3cm}
\end{figure}

In this section, we offer an intuitive justification for our proposed method, seeking to demonstrate its efficiency in extracting target domain knowledge while avoiding domain-shift accumulation. Additional details on these justifications can be found in Appendix C.

\textbf{Inter-Domain Divergence.}
To provide clearer evidence for the intuition of our proposed DaM and HOG reconstruction mechanism, we calculate the distribution distances of the feature representations across different target domains. Following prior works \cite{ruder2017learning,allaway2021adversarial, liu2023vida}, we compute the Jensen–Shannon ($JS$) divergence between two adjacent domains to represent the inter-domain divergence.
If the inter-domain divergence is small, it indicates that the feature representation remains consistent and is less susceptible to cross-domain shifts \cite{ganin2016domain}.
For comparative experiments, we compute the inter-domain $JS$ divergence of the source model, using the DaM mechanism and our proposed method on the CIFAR10-to-CIFAR10C. As illustrated in Figure \ref{fig: div}, when using the DaM mechanism, the inter-domain divergence is smaller than the source model on the vast majority of adjacent domains. Our method achieves the minimum inter-domain divergence on all fourteen adjacent domains. 
The observed pattern in inter-domain divergence suggests that DaM excels in extracting target domain knowledge but may exhibit limitations in robustness during continual adaptation. Meanwhile, after the HOG reconstruction of masked tokens, the model attains increased stability in cross-domain learning, mitigating the impact of domain shift erosion.

\begin{figure}[t]
\includegraphics[width=0.47\textwidth]{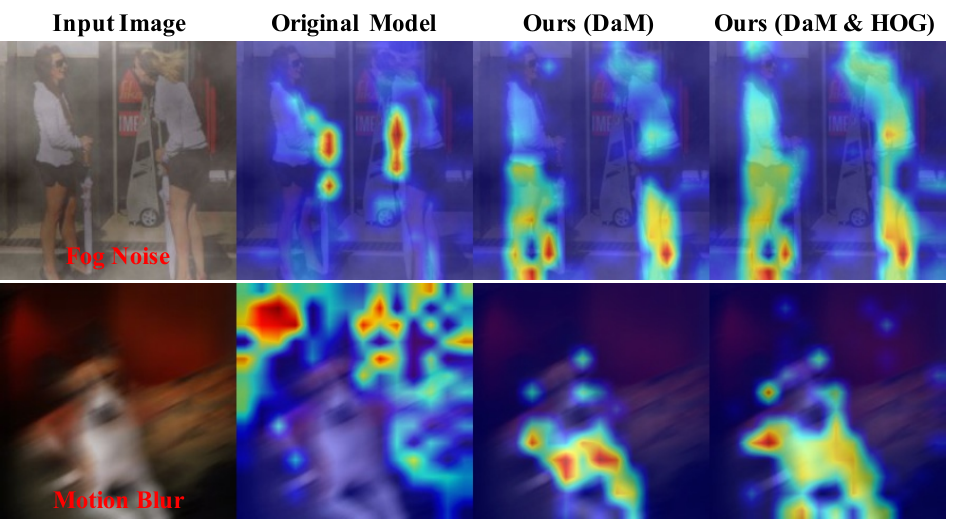}
\vspace{-0.1cm}
\centering
\caption{
The CAM visualizations.
}
\label{fig: cam}
\vspace{-0.4cm}
\end{figure}

\textbf{Class Activation Mapping (CAM).}
To directly validate our intuition, we employ CAM visualization on the ImageNet-C dataset. As shown in Figure \ref{fig: cam}, when utilizing only the source model, the attention of the features appears scattered. This dispersion is a consequence of the domain shift influence, causing the model to struggle in focusing on foreground samples. In contrast, with the DaM mechanism, there is a noticeable concentration of attention on foreground samples, indicating that DaM assists the model in better understanding the target domain knowledge. Our approach explores the domain-invariant property of HOG features. Through HOG reconstruction, we further enhance the model's task-relevant feature representations, enabling the output features to disregard background domain shift and attain higher response values on the foreground samples.

\section{Conclusion}
In this paper, we pioneer the integration of reconstruction techniques to tackle the Continual Test-Time Adaptation (CTTA) problem. Our contribution, the Adaptive Distribution Masked Autoencoders (ADMA) method, represents a novel approach to continual self-supervised learning. ADMA is designed to enhance the extraction of target domain knowledge while mitigating the accumulation of distribution shift, thereby addressing the issues of error accumulation and catastrophic forgetting. The proposed Distribution-aware Masking (DaM) mechanism plays a pivotal role in promoting the extraction of target domain knowledge, improving the efficiency of continual adaptation. Simultaneously, our introduced HOG reconstruction strategy elevates the task-relevant representations of the model and acts as a preventive measure against distribution shift accumulation.

\paragraph{Acknowledgements.} Shanghang Zhang is supported by the National Science and Technology Major Project of China (No. 2022ZD0117801).

{
    \small
    \bibliographystyle{ieeenat_fullname}
    \bibliography{main}
}
% \clearpage
\appendix

The supplementary materials presented in this paper offer a comprehensive quantitative and qualitative analysis of the proposed method. In Appendix~\ref{sec: fg}, we present the fine-grained experiment results of various methods on the ImageNet-to-ImageNet-C CTTA task scenario. The additional ablation study is provided in Appendix~\ref{sec: aas}, encompassing experiments aimed at validating the effectiveness of individual components on the ImageNet-to-ImageNet-C task. 
Meanwhile, an exploratory experiment is conducted to assess the sensitivity of the masking ratio in our proposed method. We present additional qualitative visualization comparisons on the Cityscapes-to-ACDC CTTA task in Appendix~\ref{sec: ssv}. In Appedix~\ref{sec: adj}, we furnish supplementary empirical observations and justifications supporting our motivation, including detailed computational procedures and quantitative analyses.

\section{Additional Quantitative Analysis}
\label{sec:Appendix}
\subsection{Fine-Grained Performance}
In this section, we provide a fine-grained performance of the classification results on the ImageNet-to-ImageNet-C task presented in our submissions. As shown in Table \ref{tab: imagenet-fg}, our Adaptive Distribution Masked Autoencoders (ADMA) archive the lowest classification error rate 42.5\%, and at the fine-grained level, demonstrate outstanding performance across 13 out of the 15 corruption types, validating the robustness of our method in the continual adaptation process.

\label{sec: fg}

\begin{table*}[t]
% \vspace{-0.1cm}
\centering
\small
\setlength\tabcolsep{2pt}%调列距
\begin{adjustbox}{width=1\linewidth,center=\linewidth}
\begin{tabular}{c|c|ccccccccccccccc|cc}
\toprule
% Time & \multicolumn{15}{l|}{$t\xrightarrow{\hspace*{13.5cm}}$}& \\ \hline
 Method & REF &
 \rotatebox[origin=c]{50}{Gaussian} & \rotatebox[origin=c]{50}{shot} & \rotatebox[origin=c]{50}{impulse} & \rotatebox[origin=c]{50}{defocus} & \rotatebox[origin=c]{50}{glass} & \rotatebox[origin=c]{50}{motion} & \rotatebox[origin=c]{50}{zoom} & \rotatebox[origin=c]{50}{snow} & \rotatebox[origin=c]{50}{frost} & \rotatebox[origin=c]{50}{fog}  & \rotatebox[origin=c]{50}{brightness} & \rotatebox[origin=c]{50}{contrast} & \rotatebox[origin=c]{50}{elastic\_trans} & \rotatebox[origin=c]{50}{pixelate} & \rotatebox[origin=c]{50}{jpeg}
& Mean$\downarrow$ & Gain\\\midrule
Source \cite{dosovitskiy2020image}& ICLR2021&53.0&51.8&52.1&68.5&78.8&58.5&63.3&49.9&54.2&57.7&26.4&91.4&57.5&38.0&36.2&55.8&0.0\\
Pseudo-label \cite{Leeetal2013}& ICML2013&45.2&40.4&41.6&51.3&\textbf{53.9}&45.6&47.7&40.4&45.7&93.8&98.5&99.9&99.9&98.9&99.6&61.2&-5.4\\
TENT-continual \cite{DequanWangetal2021}& ICLR2021&52.2&48.9&49.2&65.8&73&54.5&58.4&44.0&47.7&50.3&23.9&72.8&55.7&34.4&33.9&51.0&+4.8\\
CoTTA \cite{wang2022continual}& CVPR2022&52.9&51.6&51.4&68.3&78.1&57.1&62.0&48.2&52.7&55.3&25.9&90.0&56.4&36.4&35.2&54.8&+1.0\\
VDP \cite{gan2023decorate} & AAAI2023&52.7&51.6&50.1&58.1&70.2&56.1&58.1&42.1&46.1&\textbf{45.8}&23.6&70.4&54.9&34.5&36.1&50.0&+5.8\\
\textbf{Ours} & \textbf{Proposed} & \textbf{46.3} & \textbf{41.9} & \textbf{42.5} & \textbf{51.4} & 54.9 & \textbf{43.3} & \textbf{40.7} & \textbf{34.2} & \textbf{35.8} & 64.3 & \textbf{23.4} & \textbf{60.3} & \textbf{37.5} & \textbf{29.2} & \textbf{31.4} &\textbf{42.5}& +\textbf{13.3}\\
\bottomrule
\end{tabular}
\end{adjustbox}
% \vspace{-0.3cm}
\caption{\label{tab: imagenet-fg}A fine-grained Classification error rate(\%) for standard ImageNet-to-ImageNet-C online CTTA task. Mean(\%) denotes the average error rate across 15 target domains. Gain(\%) represents the percentage of improvement in model accuracy compared with the source method.}
% \vspace{-0.2cm}
\end{table*}

\subsection{Additional Ablation Study}
\label{sec: aas}

% \textbf{Effectiveness of each component on ImageNet-to-ImageNet-C}
\textbf{Components Effectiveness on ImageNet-to-ImageNet-C.} 
We conduct an additional experiment to evaluate each component of our proposed method on the ImageNet-to-ImageNet-C CTTA task. Consistent with our submission, we perform four sets of ablation studies.
As shown in Table \ref{tab: aas-imagenet}, the first set of experiments (Ex1) is applying random masking strategy to establish consistency constraints between the model outputs generated from the masked target samples and those from the original target samples. This obtains a 6.4\% reduction in the error rate in contrast to the source method (Ex0). Secondly, with the implementation of the Distribution-aware Masking (DaM) mechanism, the error rate (Ex2) is further reduced to 47.9\%, validating that DaM significantly enhances the model's ability to understand the target domain distribution. The remaining two sets of experiments (Ex3 and Ex4) are reconstructing the Histogram of Oriented Gradients (HOG) feature representations based on two masking strategies. Random masking strategy with HOG reconstruction scheme (Ex3) achieves a 1.7\% reduction in the error rate compared to use random masking strategy individually (Ex1). Our method (Ex4) outperforms others, showcasing the best results, with a remarkable 12.2\% reduction in error rate compared to the source method. These results confirm that incorporating HOG reconstruction into the continual adaptation process aids the model in acquiring task-relevant knowledge, particularly in the presence of domain shifts.

\begin{table}[t]
\centering
\setlength\tabcolsep{8pt}%调列距
% \begin{adjustbox}{width=1\linewidth,center=\linewidth}
\small
\begin{tabular}{c|ccc|c|c}
\toprule
% Time & \multicolumn{15}{l|}{$t\xrightarrow{\hspace*{13.5cm}}$}& \\ \hline
 &Random& DaM &HOG
& Mean$\downarrow$ &Gain \\\midrule
Ex0& - & - & - &55.8&/\\
Ex1& \checkmark & -&- &49.4& +6.4\\
Ex2& - &\checkmark &- &47.9& +7.9\\
Ex3& \checkmark & -&\checkmark &47.7& +8.1\\
Ex4&-  & \checkmark&\checkmark &\textbf{43.6} & +\textbf{12.2} \\
\bottomrule
\end{tabular}
% \end{adjustbox}
% \vspace{-0.3cm}
\caption{Average error rate(\%) for ImageNet-to-ImageNet-C online CTTA task. Random, DaM, and HOG represent the random masking strategy, our proposed Distribution-Aware Masking mechanism, and our introduced HOG reconstruction, respectively.}
% \vspace{-0.1cm}
\label{tab: aas-imagenet}
\end{table}

\begin{figure}[t]
\includegraphics[width=0.47\textwidth]{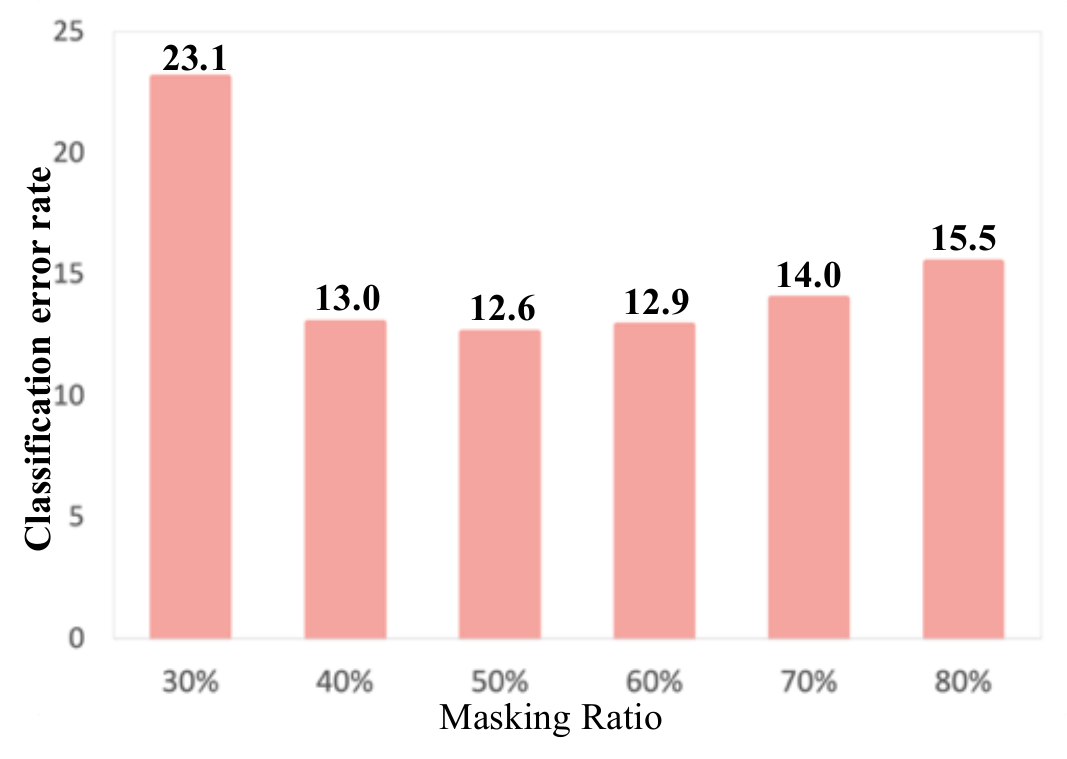}
% \vspace{-0.1cm}
\centering
\caption{
Average error rate(\%) for CIFAR10-to-CIFAR10C CTTA task when applying different masking ratio(\%).
}
\label{fig: maskratio}
% \vspace{-0.3cm}
\end{figure}

\textbf{Masking Ratio Sensitivity.} 
We conduct another set of ablation experiments to investigate the sensitivity of our DaM mechanism to the masking ratio. Given that the optimal results are obtained when DaM is coupled with the reconstructed HOG scheme, we specifically conduct these experiments directly on the CIFAR10-to-CIFAR10C CTTA task using our integrated method (DaM\&HOG).
The effect of the masking ratio is shown in Figure \ref{fig: maskratio}, a wide range of mask ratios from 30\% to 80\% produce different performances.  
Clearly, the optimal result showcased in the submission is achieved when the masking ratio is set at 50\%. Remarkably, within the vicinity of a 50\% masking ratio, the error rate exhibits minimal fluctuations, with the highest recorded error rate being 14.0\% at a 70\% masking ratio.
However, setting the masking ratio to the extremes (i.e., 30\% and 80\%), yields intriguing results. 
The large masking ratio of 80\% results in a 2.9\% deterioration in the error rate compared to the best result, while the small masking ratio of 30\% leads to a more pronounced degradation of 10.5\%.
Hence, the results suggest that DAM is not highly sensitive to the masking ratio when it exceeds 30\%, consistently yielding a relatively robust adaptation process.
At the same time, when the masking ratio is too large, such as 70\% or 80\%, the classification error rate also experiences some increase. The reason is that the mask covers a large portion of information, resulting in insufficient semantic information expression. Finally, we chose a masking ratio of 50\% for our classification and segmentation CTTA experiments.

\begin{figure*}[htb]
\centering
\includegraphics[width=\linewidth]{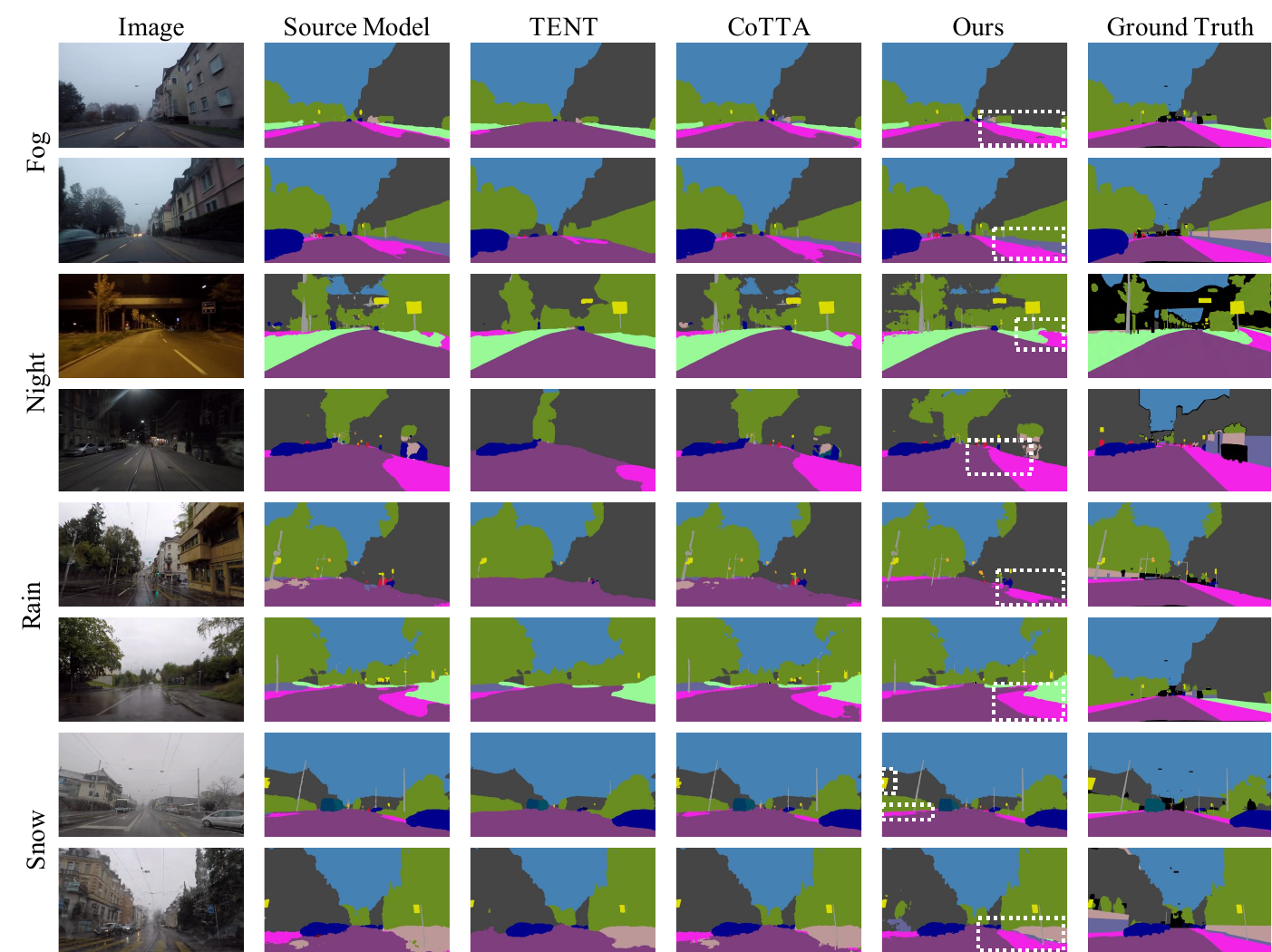}
\caption{Qualitative comparison of our method with previous SOTA methods on the ACDC dataset. Our method could better segment different pixel-wise classes such as shown in the white box.}
\label{fig:vis_sup}
\end{figure*}

\section{Additional Qualitative Analysis}
\label{sec: ssv}

To intuitively assess the effectiveness of our approach, we conducted an additional set of qualitative experiments. Specifically, we performed four sets of comparative experiments in the Cityscapes-to-ACDC CTTA scenario.
In the first set of experiments, we tested the Segformer-B5 model \cite{xie2021segformer}, pre-trained on the source domain Cityscapes dataset, directly on the four shifted domains of the ACDC dataset. Next, we adapted the model, initially pre-trained on the source domain, to the target domains using the leading CTTA methods TENT \cite{DequanWangetal2021} and CoTTA \cite{wang2022continual} from recent years. The final set of experiments involved applying our proposed method for continual adaptation to the four target domains.
The results of the visualization of the segmentation maps for all the methods are shown in Figure \ref{fig:vis_sup}. The model applying our method has the best segmentation results for all the target domains compared to the original source model, the model applying the TENT method and the model applying the CoTTA method. Notably. Our method archives consistent improvements for most categories and the benefits in categories like $sidewalk$, $terrain$, and $traffic\ sign$ are very significant (shown in white box). 
Adapting to these challenging categories is inherently difficult. Therefore, our DaM mechanism strategically masks samples from these categories that are more susceptible to domain shifts during the testing process. Simultaneously, our model continues to process inputs from the original images. By leveraging this contextual knowledge for consistency constraints, coupled with the HOG reconstruction scheme that enhances task-relevant feature representation, the model achieves more accurate segmentation of intricate regions.

\section{Additional Discussion and Justification}
\label{sec: adj}

In this section, our goal is to furnish detailed implementation insights that substantiate our intuition.
In Section \ref{absec: div}, we elucidate our choice of utilizing the Jensen–Shannon (JS) divergence for computing inter-domain divergence. 
Additionally, we illustrate the trends of inter-domain divergence in segmentation tasks.
We extend the visualization of Class Activation Mapping (CAM) in Section \ref{absec: cam}.

\subsection{Inter-Domain Divergence.}
\label{absec: div}
To substantiate the rationale behind our proposed DaM and HOG reconstruction mechanism, we measure the distribution distances of feature representations across various target domains.
Inspired by previous works \cite{ganin2015unsupervised}, we utilize the domain distance definition proposed by Ben-David \cite{ben2006analysis} and employ the $\mathcal{H}$-divergence metric to assess the domain representations of our proposed method. The $\mathcal{H}$-divergence between $D_S$ and $D_{T_i}$ can be calculated as:
\begin{equation}
\begin{split}
d_\mathcal{H}(D_S, D_{T_i}) = 2 \mathop{\mathrm{sup}}_{\mathcal{D} \sim \mathcal{H}} | \mathop{\mathrm{Pr}}_{x \sim D_S}[\mathcal{D}(x)=1] - \\\mathop{\mathrm{Pr}}_{x \sim D_{T_i}}[\mathcal{D}(x)=1] |
\end{split}
\label{eq:dh}
\end{equation}
, where $\mathcal{H}$ denotes hypothetical space and $\mathcal{D}$ denotes discriminator. 
Drawing inspiration from \cite{ruder2017learning,allaway2021adversarial, liu2023vida}, we employ the Jensen–Shannon (JS) divergence between two adjacent domains as an approximation of $\mathcal{H}$-divergence, as it has proven effective in distinguishing domain representations. When the inter-domain divergence is relatively small, it indicates a consistent feature representation less affected by cross-domain shifts \cite{ganin2016domain}.
\begin{equation}
\begin{split}
JS(P_{D_S} || P_{D_{T_i}}) = \frac{1}{2}KL(P_{D_S}|| \frac{P_{D_S}+P_{D_{T_i}}}{2}) + \\ \frac{1}{2}KL(P_{D_{T_i}}|| \frac{P_{D_S}+P_{D_{T_i}}}{2})
\label{eq:js}
\end{split}
\end{equation}
Where $Kullback$-$Leibler\ (KL)\ divergence$ between two domain is

\begin{equation}
KL(P_1||P_2) = \sum_{i=0}^{n} P_1(x_i) log(\frac{P_1(x_i)}{P_2(x_i)})
\label{eq:kl}
\end{equation}
Utilizing $P$ to represent the probability distribution of the model output features, we partition the output feature space into mutually disjoint intervals $x_i$, where $n$ denotes the total number of samples in each target domain. As depicted in Figure 4 of the main paper, our proposed method exhibits a gradual reduction in inter-domain divergence.

\begin{figure}[t]
\includegraphics[width=0.47\textwidth]{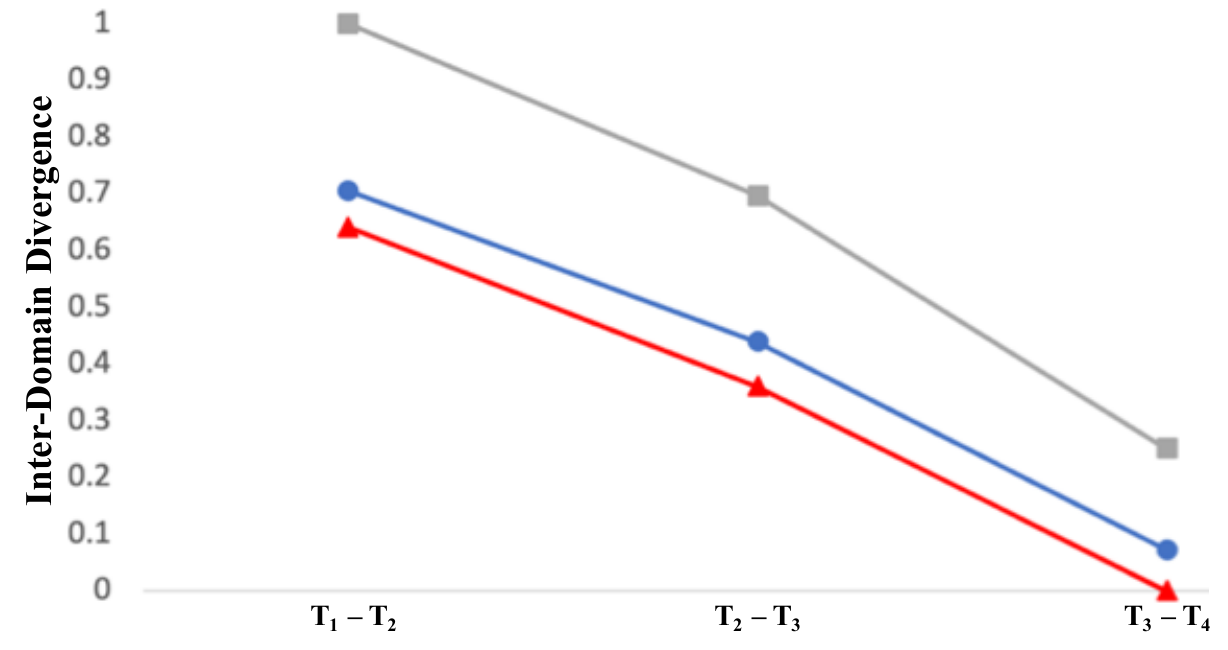}
% \vspace{-0.1cm}
\centering
\caption{
The inter-domain divergency. $T_{1}$ to $T_{4}$ represent the 4 target domains in ACDC, listed in sequential order.
}
\label{fig: segdiv}
% \vspace{-0.3cm}
\end{figure}

Furthermore, we apply the same approach to calculate inter-domain divergence in the segmentation CTTA task, conducted on the Cityscapes-to-ACDC scenario. As depicted in Figure \ref{fig: segdiv}, the DaM mechanism yields smaller inter-domain divergence compared to the source model across all adjacent domains. Our method further reduces the divergence on all adjacent domains. The results demonstrate that DaM excels in extracting target domain knowledge, while HOG reconstruction increases stability in cross-domain learning and mitigates the impact of domain shift erosion.

\begin{figure}[htp]
\includegraphics[width=0.47\textwidth]{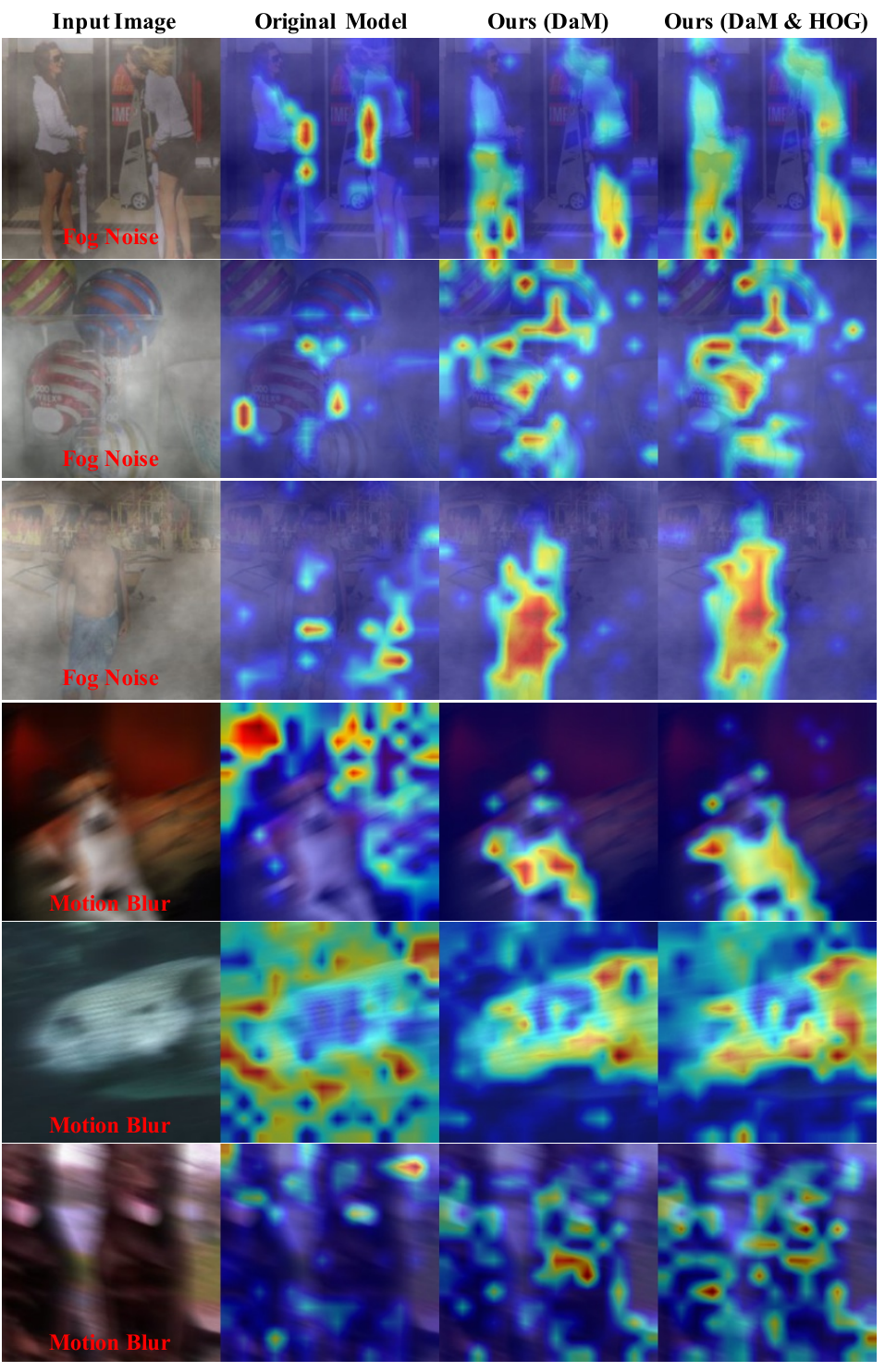}
% \vspace{-0.1cm}
\centering
\caption{
The CAM visualizations.
}
\label{fig: supcam}
% \vspace{-0.4cm}
\end{figure}

\subsection{Class Activation Mapping (CAM)}
\label{absec: cam}
To empirically validate our intuition, we extend the use of CAM visualization to a larger set of samples within the ImageNet-C dataset. As illustrated in Figure \ref{fig: supcam}, the results demonstrate a consistent trend with those presented in the submission. Specifically, when employing only the source model, the attention of the features appears scattered. This dispersion is a consequence of the domain shift influence, which hinders the model's ability to focus on foreground samples. In contrast, with the DaM mechanism, there is a noticeable concentration of attention on foreground samples, indicating that DaM assists the model in better understanding the target domain knowledge. Our approach leverages the domain-invariant property of HOG features. Through HOG reconstruction, we further enhance the model's task-relevant feature representations, allowing the output features to disregard background domain shift and achieve higher response values on the foreground samples.


\end{document}